# CWoMP: Morpheme Representation Learning for Interlinear Glossing

**Morris Alper**[*1,2] **Enora Rice**[*3] **Bhargav Shandilya**[*3] **Alexis Palmer**[3] **Lori Levin**[1]

[*]Equal contribution [1]Carnegie Mellon University [2]University of Miami [3]University of Colorado Boulder

https://cwomp.github.io

## Abstract

Interlinear glossed text (IGT) is a standard notation for language documentation which is linguistically rich but laborious to produce manually. Recent automated IGT methods treat glosses as character sequences, neglecting their compositional structure. We propose *CWoMP (Contrastive Word-Morpheme Pretraining)*, which instead treats morphemes as atomic form-meaning units with learned representations. A contrastively trained encoder aligns words-in-context with their constituent morphemes in a shared embedding space; an autoregressive decoder then generates the morpheme sequence by retrieving entries from a mutable lexicon of these embeddings. Predictions are interpretable–grounded in lexicon entries–and users can improve results at inference time by expanding the lexicon without retraining. We evaluate on diverse low-resource languages, showing that CWoMP outperforms existing methods while being significantly more efficient, with particularly strong gains in extremely low-resource settings.

## 1 Introduction

While thousands of languages are spoken in the world, the majority are at risk of *language loss*—ceasing to be passed on to future generations (Pine and Turin, 2017). As global language diversity dwindles, language documentation is increasingly important to preserve minority languages, aid revitalization efforts, and keep a scientific record for posterity (Seifart et al., 2018).

A common form of linguistic annotation used in language documentation is *interlinear glossed text (IGT)*, a format which illustrates the internal structure of words and encodes the functional and lexical meanings of their constituent morphemes. In this work, we define each such morpheme as a *paired segment and gloss*—that is, a *form-meaning unit* (see Section 2).

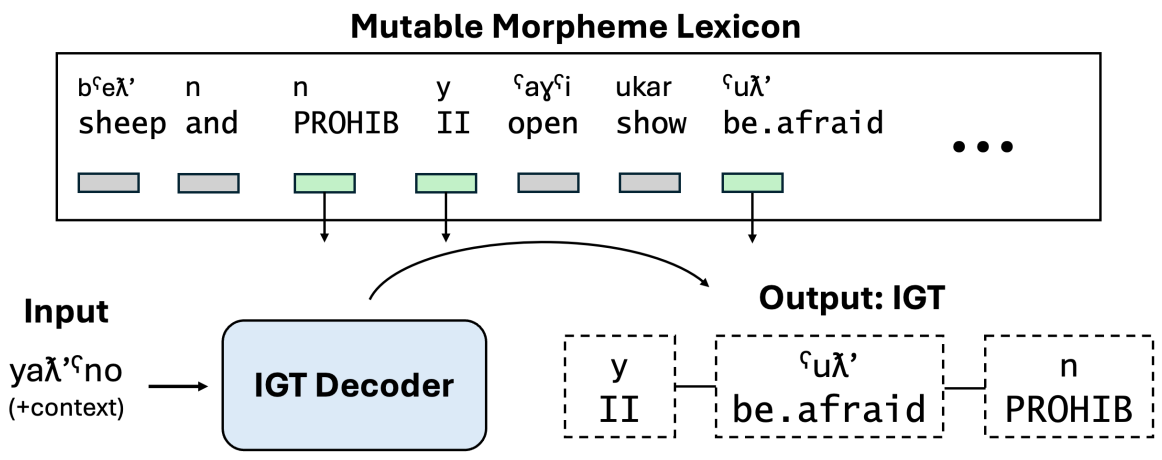

Figure 1: CWoMP predicts key components of IGT by retrieving morpheme entries from a mutable lexicon of learned representations (green morpheme embeddings are selected). Compared to approaches that generate unconstrained text, our method's outputs are interpretable and can be refined at inference time by adding lexicon entries without retraining. In the example shown above, a Tsez word meaning *Don't be afraid* is split by CWoMP into constituent morphemes.

Though IGT is widely used in language documentation, producing it is time-consuming, motivating growing interest in automated glossing (Ginn et al., 2024c). So far, existing systems have seen limited adoption due to a disconnect between research output and practical needs of field linguists (Gessler, 2022; Rice et al., 2025). To address some of these needs, we propose a novel system, *CWoMP* (*Contrastive Word-Morpheme Pretraining*, see Figure 1). CWoMP takes unsegmented transcriptions as input and jointly predicts morphological segmentation and glosses. A contrastively trained encoder aligns words and their constituent units in a shared embedding space, producing a discrete codebook of morpheme segments and their glosses. An autoregressive decoder then generates the morpheme sequence for each target word by retrieving entries from this codebook.

This system offers several unique advantages: (A) Because predictions are constrained to codebook entries, they are *free of hallucinated morpheme types* by design. (B) Each entry contains both a segment and a gloss, guaranteeing that the number of glosses is equal to the number of morphemic seg-ments. (C) The codebook is *extensible at inference time*: users can add new morphemes to handle unseen vocabulary without retraining. (D) Predictions are *interpretable* as they are grounded in a known lexicon and their embeddings reflect morphology via geometric relations. (E) By using meaningful representations for morphology, our method is significantly more data- and compute-efficient than prior SOTA.

Our experiments on seven typologically diverse low-resource languages (Section 4.1) show that CWoMP outperforms existing SOTAs at automatic IGT generation, with particularly large gains in extremely low-resource settings, the scenario most relevant for field linguists. Performance consistently improves when the codebook is expanded with ground-truth morphemes at inference time. This is an oracle-style upper bound (Section 4.2) rather than a fully automatic result, but it approximates the practical workflow of a linguist refining outputs without retraining. CWoMP also achieves these results with significantly lower computational cost than prior approaches. We release our code and trained models at our project page.[1]

## 2 Background – Interlinear Glossed text

IGT is a structured linguistic annotation format that aligns text in a target language to labels describing the meanings and grammatical properties of individual morphemes. A typical IGT instance consists of a transcription, segmentation, gloss, and translation, as in the following Tsez example (Abdulaev and Abdullaev, 2010):

| | | | | | | | |
|---|---|---|---|---|---|---|---|
| (1) | T’ay | rił | łu | ragäƛin | | | |
| | t’ay | rił | łu | r- | agi | -a | -ƛin |
| | from.here | butter | who.ERG | IV- | lick | -Q | -QUOT |

“Who licked the butter out of it?”

The second line (segmentation tier) segments words at morpheme boundaries (marked by hyphens). The third line (gloss tier) glosses each segment: grammatical labels appear in capitals (e.g., ERG), lexical meanings in lowercase (e.g., ‘butter’), and periods indicate fusional morphemes carrying multiple functions. These conventions follow the Leipzig glossing rules (Lehmann, 1982).

We note that there are multiple conventions for segmentation due to the fact that morphemes often change form in context—e.g., English *leaf* with the plural suffix *-s* becomes *leaves*. This could be segmented as *leave-s* (surface segmentation, preserving the form as written) or *leaf-s* (underlying segmentation, also called canonical segmentation). In our results (Section 4.3) we discuss how segmentation type may have downstream effects on our method.

Beyond analyzing individual sentences, a collection of IGT preserves a systematic record of a language, encoding (e.g.) morphotactics, morphosyntactic strategies, and vocabulary. Some other morphological labeling schemas, like UniMorph (Batsuren et al., 2022) and Universal Dependencies (de Marneffe et al., 2021), map surface forms to a fixed, crosslingual ontology, supporting crosslingual analysis. IGT instead supports language-specific documentation and analysis, and language-specific schemas allow for annotation of idiosyncratic grammatical phenomena that are difficult to represent with standardized taxonomies. More importantly, we focus on IGT because it is the standard format for morphological annotation in language documentation and linguistic scholarship more broadly.

Automatic IGT generation must address several challenges that motivate our method. *Allomorphy:* morphemes change form in context, requiring flexible mappings between surface text and underlying forms. *Out-of-Vocabulary (OOV) morphemes:* training data rarely covers the full inventory of morphemes encountered in practice. *Hallucination:* spurious morpheme types are especially harmful for endangered languages, where uncaught errors may propagate. *Contextual disambiguation:* homographs like English *bears* may require different analyses (‘bear’-PL vs. ‘bear’-3SG) depending on context. Finally, documentary linguists need interpretable, refinable predictions that work in extremely low-resource settings. CWoMP is designed to address each of these challenges, as outlined in Section 1.

## 3 Method

Here, we introduce the CWoMP system for gloss generation. We begin with preliminary definitions and prompt formats (Section 3.1), then describe the two stages of CWoMP: contrastive pretraining of morpheme representations (Section 3.2), followed by autoregressive decoding over the resulting lexicon (Section 3.3). Further implementation details are provided in Appendix A.1.

### 3.1 Preliminaries

Let $\mathcal{W}$ denote the vocabulary of **words-in-context**: structured inputs that couple each target word with a transcription of the context in which it appears and the corresponding translation (see Example 2). Let $\mathcal{M}$ denote the candidate pool of **morphemes**. Each word $w \in \mathcal{W}$ corresponds to ground truth IGT representation $G_w = (m_1, m_2, \cdots, m_k)$ where each $m_i = (s_i, g_i) \in \mathcal{M}$ is a morpheme with segment $s_i$ and gloss $g_i$ (both strings). Note that $w$ and $s_i$ are both in the target language, while $g_i$ is gloss text in a language of wider communication, including grammatical labels. We also define the *bag-of-morphemes* (BoM) representation of a word $w$ as the unordered set $B_w = \{m_1, \cdots, m_k\}$.

[1] `https://cwomp.github.io`

A concrete example of these definitions is given below, using a word in the context of Example 1:

$$w = \text{ragäƛin (+ context)}$$

$$G_w = (\underbrace{(\text{r}, \text{IV})}_{m_1}, (\underbrace{\text{agi}}_{s_2}, \underbrace{\text{lick}}_{g_2}), (\text{a}, \text{Q}), (\text{ƛin}, \text{QUOT}))$$

$$B_w = \{m_1, m_2, m_3, m_4\} \quad \text{(unordered)}$$

In practice, words-in-context and morphemes are represented by prompt strings that are converted into embeddings by the BoM encoder. Schematic examples of these strings are shown below:

(2) $w$: "ragäƛin | Context: T'ay riɬ ɬu ragäƛin? | Translation: Who licked the butter out of it?"
$m$: "ƛin | Gloss: QUOT"

Text with dotted underlines has spaces inserted between characters—forcing character-level tokenization to avoid subword tokenization artifacts. We include sentential context which may benefit glossing (Zhao et al., 2020; Ginn et al., 2023); the target word is always isolated in its own field of the prompt, while inter-character spacing within the context field does not preserve word boundaries. For languages with provided translations, this is concatenated to the prompt as an additional field, as seen above. We ablate these design choices in Appendix B.4.

Words that occur multiple times in the training data will yield multiple distinct training instances. For example, the Natugu word *ma* ('house') occurring in the sentences *Ma yrkrtxo-kz* and *sc tqvzpebo ma nyzde* yields two unique training instances with different context fields containing these respective sentences.[2]

### 3.2 BoM Encoder

In order to learn meaningful representations of words and morphemes, we use an inclusion objective: Given word-in-context $w$ and morpheme $m$, the goal is to determine whether $m \in B_w$, where $B_w$ is the BoM representation described above.

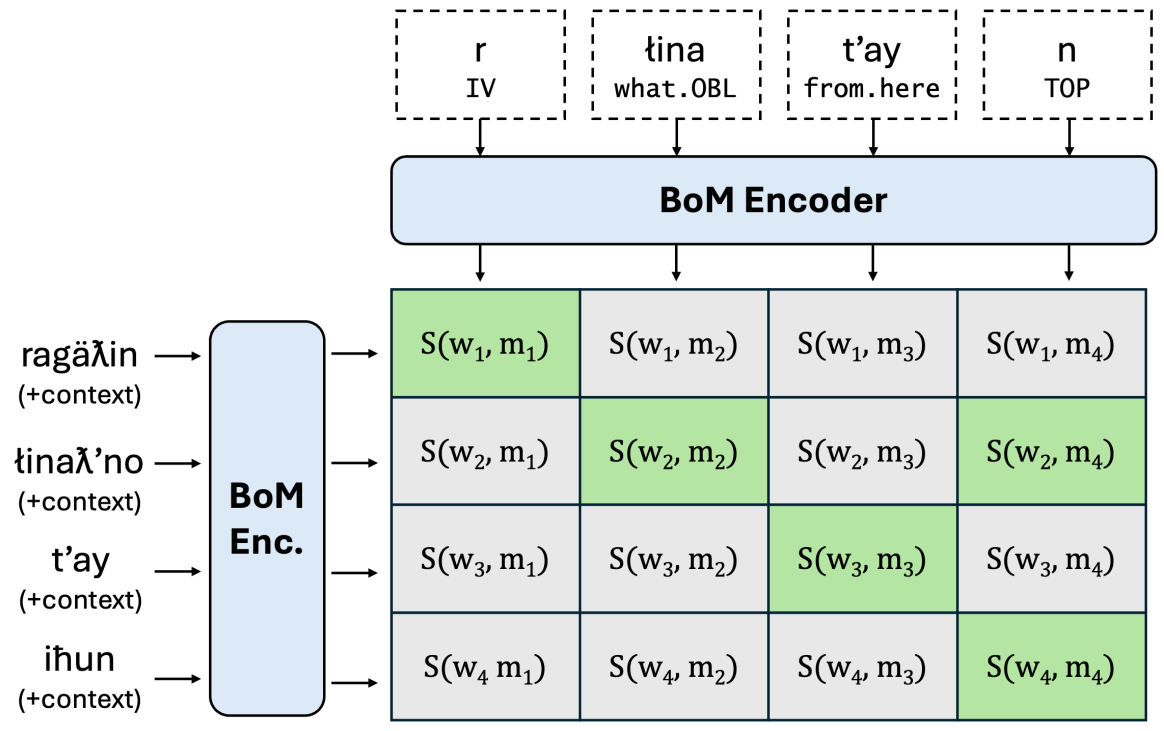

Figure 2: **BoM Encoder.** This dual encoder learns to embed words and morphemes in a shared space via contrastive learning. The distance between the embedding of a word $w$ and the embeddings of its constituent morphemes (shown in green) is minimized, while the distance between $w$ and all other morphemes is maximized. Above, $S(w_i, m_j)$ denotes the cosine similarity between encodings of the $i$-th word and $j$-th morpheme.

We employ a dual encoder architecture (Figure 2) that learns embedding functions $p_\theta : \mathcal{W} \to \mathbb{R}^n$ and $q_\theta : \mathcal{M} \to \mathbb{R}^n$, mapping words-in-context and morphemes into a shared latent space. Both use a single string encoder—a transformer with shared weights $\theta$, initialized from a pretrained multilingual text encoder. These encoders apply L2 normalization to their outputs, and the similarity between a word-in-context and morpheme is given by their dot-product similarity $S(w, m) = p_\theta(w) \cdot q_\theta(m)$. Intuitively, this similarity should be high for morphemes contained in the target word and low otherwise.

We train with a multi-positive variant of the InfoNCE loss (Oord et al., 2018) using in-batch negatives, related to the supervised contrastive loss of Khosla et al. (2020). For a batch of $B$ word–morpheme pairs $\{(w_i, m_i)\}_{i=1}^{B}$, each sample pairs a word-in-context with one of its constituent morphemes (randomly chosen), so $(w_i, m_i)$ are positive matches for each $i$. In standard InfoNCE, all other in-batch morphemes are treated as implicit negatives. However, since words share morphemes—e.g., the same grammatical affix may appear in multiple words—other in-batch morphemes may also be valid positives for a given query. Naively treating these as negatives penalizes the model for correctly recognizing shared morphemes (hurting performance; see Appendix B.4). We address this by identifying all valid positive pairs within each batch and averaging over them, normalizing per query to prevent those with more in-batch positives from dominating the gradient. The loss is thus given by

$$\mathcal{L}(\theta) = \frac{-1}{B}\sum_{i=1}^{B}\frac{1}{|\mathcal{P}_i|}\sum_{p\in\mathcal{P}_i}\log\frac{e^{S(w_i,m_p)/\tau}}{\sum_{j=1}^{B}e^{S(w_i,m_j)/\tau}},$$

where $\mathcal{P}_i$ is the set of valid in-batch positives for query $w_i$ and $\tau$ is a learnable temperature parameter. When $|\mathcal{P}_i| = 1$ this reduces to standard InfoNCE.

[2] Note that *r*, *x*, *c*, *q*, and *z* represent vowel sounds in Natugu orthography.

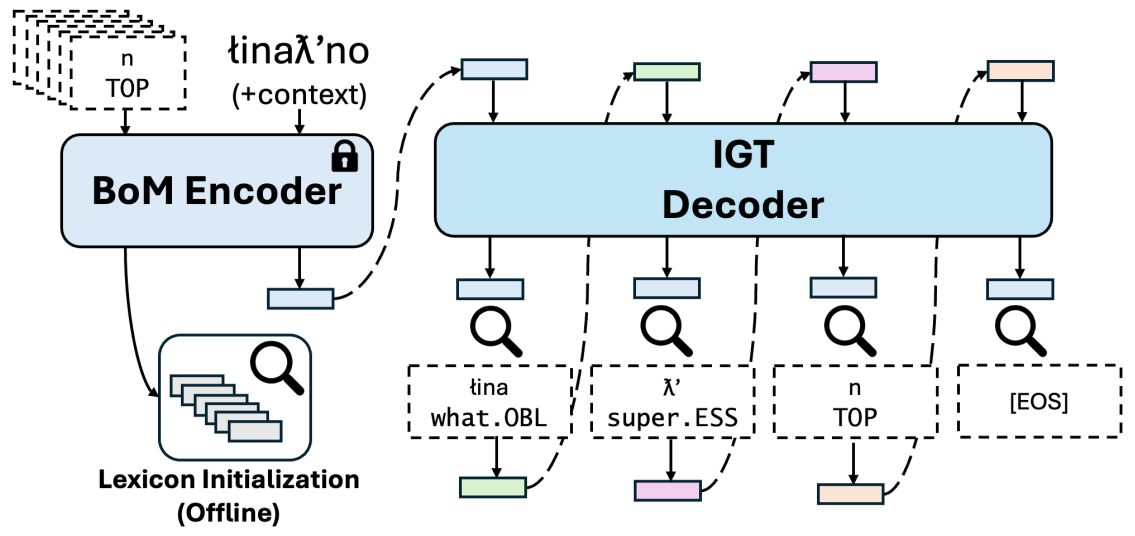

Figure 3: **IGT Decoder.** To predict IGT, CWoMP autoregressively predicts morphemes as form-meaning units, conditioned on an input word. The pretrained BoM encoder is frozen (lock symbol) and used offline to cache embeddings of all lexicon morphemes. During inference, the word-in-context to be glossed is encoded and passed autoregressively through the IGT decoder; at each step, the latter outputs an embedding which is matched (magnifying glass symbol) to its nearest neighbor in the lexicon, whose encoding is fed back into the decoder.

### 3.3 IGT Decoder

Given a query word-in-context $w$, the decoder (Figure 3) autoregressively generates a sequence of embeddings, each matched to its nearest neighbor in a precomputed lexicon of morpheme embeddings produced by the frozen BoM encoder. The codebook contains entries for all morphemes observed during training, plus a special [EOS] token. Each unique (segment, gloss) pair constitutes a distinct codebook entry with its own embedding; homographic segments with multiple potential glosses (e.g., Tsez *-n* TOP vs. *-n* 'and') thus yield separate entries, which the decoder disambiguates using sentential context. Importantly, this means the decoder is constrained to select from these morpheme entries and cannot hallucinate unseen morpheme types. The decoder is run independently per word and generates the within-word morpheme sequence (e.g., stem+affixes), with sentential context provided as conditioning information via the input word representation.

The decoder is a lightweight autoregressive transformer. Its input consists of the word-in-context embedding $q_\theta(w)$ (including sentential context) followed by the embeddings of previously generated morphemes – ground truth during training (teacher forcing), or the model's own retrieved predictions at inference. Concretely, $q_\theta(w)$ occupies position 0, followed by the [BOS] token and morpheme sequence in order; the model predicts each subsequent codebook index, ending in [EOS]. Inputs are projected to the model's hidden dimension $d_{\text{model}}$ with sinusoidal positional encoding.

At each decoding step $j$, the model's hidden state $h_j$ is projected to $d_{\text{model}}$ and scored against all codebook entries via dot-product similarity:

$$\text{logits}_j = h_j \cdot V_{\text{emb}}^{\top}/\kappa$$

where $V_{\text{emb}}$ is the matrix of precomputed morpheme embeddings and $\kappa$ is a temperature parameter learned jointly with the decoder. These logits are used for standard beam search decoding.

## 4 Experiments and Results

We first discuss our experimental setup, including baselines and evaluation methodology (Sections 4.1 and 4.2). We then present our results on gloss generation (Section 4.3) and segmentation (Section 4.4). Finally, we analyze our method's data and compute efficiency (Sections 4.5 and 4.6) and interpretability of learned embeddings (Section 4.7). Further details and results including ablations and morpheme retrieval evaluation are provided in Appendix B.1.

### 4.1 Experimental Setup

**Architecture and Training.** For our dual encoder model, we initialize a pretrained LaBSE (Feng et al., 2022) transformer model, using shared weights for the word and morpheme encoders. Our selection of this backbone is motivated by its strength as a multilingual sentence encoder, confirmed by our ablation study comparing it to a comparable alternative (Appendix B.4). Our autoregressive decoder is composed of four transformer blocks and is randomly initialized.

**Data.** Data comes from the SIGMORPHON 2023 shared task on IGT Interlinear Glossing (Ginn, 2023), covering seven diverse languages: Gitksan, Lezgi, Natugu, Nyangbo, Tsez, Uspanteko, and Arapaho. All languages include translations except for Nyangbo; these are in Spanish for Uspanteko and English for other languages. Results are calculated on each language's respective test set. Dataset sizes are reported in Table 1.

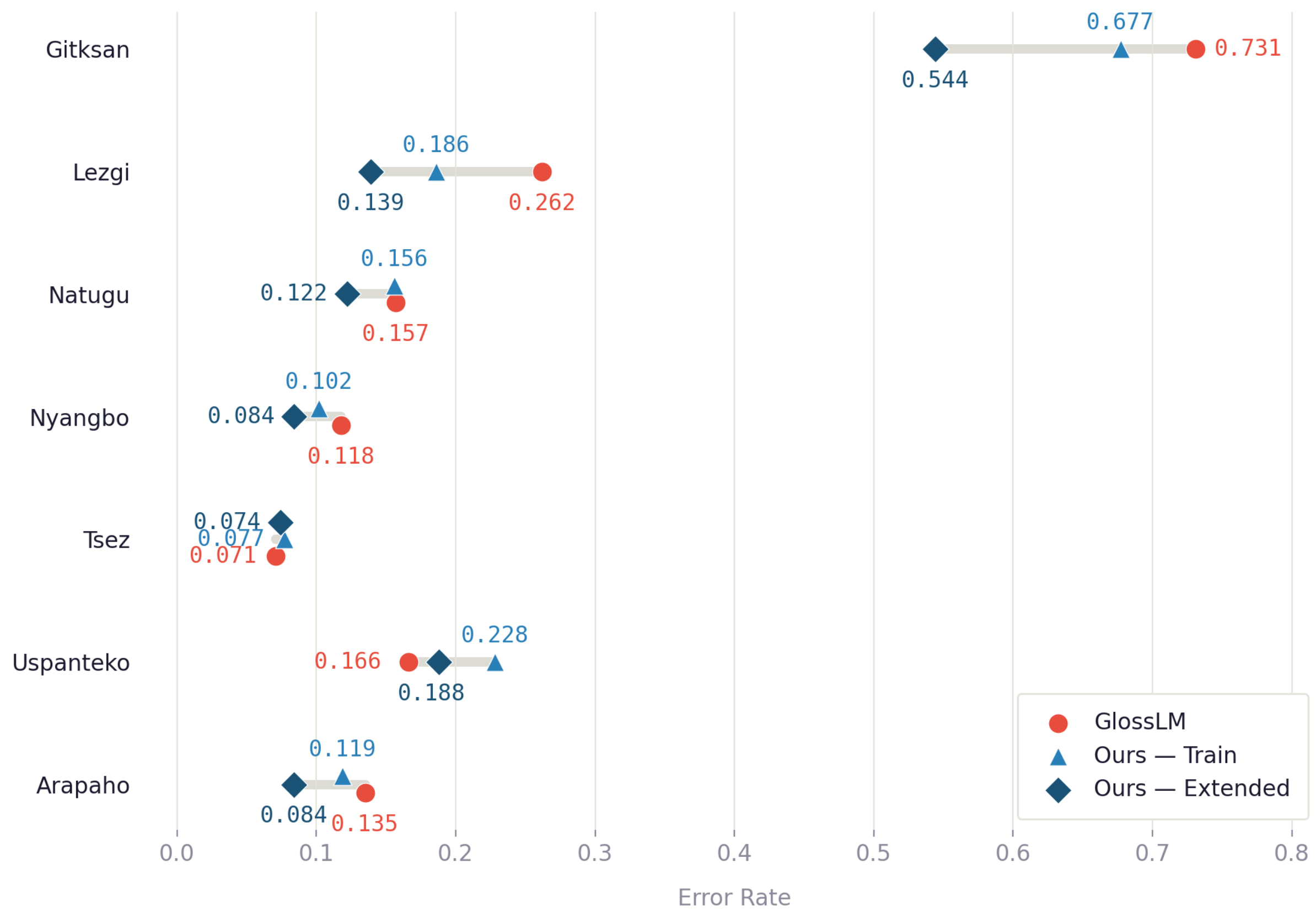

Figure 4: **Improvement over baseline**. We report Morpheme Error Rate (↓) for **glossing** for CWoMP (ours) and the leading prior method GlossLM. CWoMP performs competitively or outperforms GlossLM for most languages, with a substantial boost in the extended lexicon setting. We analyze factors impacting Uspanteko performance in Section 4.3.

| Language | Train | Dev | Test |
|---|---|---|---|
| Gitksan (git) | 31 | 42 | 37 |
| Lezgi (lez) | 701 | 88 | 87 |
| Natugu (ntu) | 791 | 99 | 99 |
| Nyangbo (nyb) | 2,100 | 263 | 263 |
| Tsez (ddo) | 3,558 | 445 | 445 |
| Uspanteko (usp) | 9,774 | 232 | 633 |
| Arapaho (arp) | 39,501 | 4,938 | 4,892 |

Table 1: **Dataset Size Statistics.** Listed values are counts of glossed sentences.

**Baselines.** We compare to the leading model for gloss generation, GlossLM (Ginn et al., 2024c), which treats gloss generation as a character-level sequence-to-sequence task while benefiting from cross-lingual transfer learning across many languages. CWoMP, by contrast, is trained separately for each target language using only that language's monolingual IGT data, without any cross-lingual pretraining (Section 4.3). Specifically, we compare our model to $\text{GlossLM}_{\text{UNSEG, FT}}$, a setting where GlossLM is trained on unsegmented input and fine-tuned on task data in the target language.

GlossLM is not trained with explicit segmentation supervision, despite the presence of a segmentation tier in the SIGMORPHON data. Although there is a version of GlossLM ($\text{GlossLM}_{\text{ALL}}$) that predicts the gloss tier from pre-segmented inputs, this task is substantially less challenging than ours. It also fails to map onto realistic documentation workflows given that linguists will typically segment and gloss in parallel and do not often produce corpora that are fully segmented but remain unglossed (Rice et al., 2025). Unlike these baselines, CWoMP is capable of effectively using both segmentation and gloss tiers as training supervision, while during inference, it receives *only* an unsegmented transcription and simultaneously predicts *both* the segmentation and the gloss. In Appendix B.3., we also provide a comparison to the concurrent glossing model of Ginn et al. (2026), whose reported results use a different data split.

**Metrics.** Our primary evaluation metric is *Morpheme Error Rate (MER)*. Following Ginn et al. (2026), this is defined as follows: given respec-tive ground truth and predicted glosses $G = (g_1, \cdots, g_n)$ and $G' = (g'_1, \cdots, g'_m)$ of an entire utterance, MER is calculated as the Levenshtein distance between these sequences, treating each gloss symbol as an atomic unit. Thus MER is equal to 0 if and only if $G = G'$, and reflects insertion, deletion, and substitution of gloss symbols, analogous to word and character error rate metrics standard for tasks such as automatic speech recognition and machine translation. We calculate MER identically for the segmentation tier, treating segments rather than gloss symbols as atomic units. We adopt MER over metrics used in prior work for several conceptual reasons: unlike the metrics of Ginn et al. (2024c), it is not biased towards the left-hand side of sentences—those metrics treat all content after the first misaligned morpheme as incorrect; unlike the F1 metric of Mager et al. (2020), MER is order-aware, and unlike word accuracy, MER gives a fine-grained signal when a word's analysis is partially correct. Moreover, MER can be applied consistently between the glossing and segmentation tasks, and signals both word- and morpheme-level performance. We report additional metrics used in prior works, including a direct empirical comparison confirming these metrics agree in relative model ranking, in Appendix B.2.

### 4.2 Mutable Lexicon Evaluation

In a realistic linguistic annotation scenario, a user may wish to expand the lexicon of recognized morphemes at inference time without additional model training. This is particularly pressing when considering field linguist users who may not be experts in model training but desire an interpretable method for improving model outputs. The mutable morpheme lexicon of CWoMP allows us to propose a new framework for testing this setting by expanding the lexicon to include ground truth morphemes unseen during training.

In particular, for all experiments we report metrics in two settings:

1. *Train Lexicon:* The set of retrievable morphemes consists only of those seen during training. OOV morphemes are matched to their nearest neighbor in the training lexicon, which may result in incorrect substitutions.

2. *Extended Lexicon:* Ground truth morphemes from the evaluation split (dev or test) are also included in the set over which retrieval is performed.

The first setting strictly precludes data leakage from evaluation sets, while the second setting approximates results in a dynamic setting with a user noticing that a certain morpheme is never predicted and adding it to the lexicon without retraining.

Note that the extended setting adds only morpheme entries to the lexicon; model weights are unchanged, and no training signal is derived from evaluation data. While this is an oracle setting that assumes knowledge of which morphemes are missing, it simulates the realistic workflow of a linguist manually expanding a lexicon and is analogous to adding entries to a dictionary used by a retrieval system.

As part of the intended human-in-the-loop workflow underlying this evaluation: a linguist reviews CWoMP's predictions, and accepted or corrected segment–gloss pairs are added to the mutable lexicon for use on subsequent sentences, without retraining. This workflow is not simulated here but is directly implemented in GLOSSASSIST (Shandilya et al., 2026), an annotation interface built around CWoMP that logs annotator accept/reject/edit decisions and feeds accepted entries back into the lexicon in real time.

### 4.3 Gloss Generation

**Task Description.** In this task, we predict glosses from sentence transcriptions and translations, reporting the metrics defined in Section 4.1.

**Results.** The results of our final configuration and corresponding baselines on glossing are shown in Figure 4, with languages sorted in ascending order of training data quantity (values in Section 4.1). These results show that CWoMP mostly achieves competitive or superior performance across languages, with consistent improvements in the extended lexicon setting. Gains are particularly pronounced for mid- to low-resource languages: on Lezgi, CWoMP reduces MER from 0.26 to 0.14–0.19 (extended vs. train vocab), and on Gitksan–with only 31 training examples–the extended vocabulary setting reduces MER from 0.68 to 0.54. As discussed in Section 4.2, these extended-lexicon numbers reflect an oracle upper bound rather than a fully automatic result, since they assume the missing ground-truth morphemes are already known and added to the lexicon. CWoMP also improves performance on the highest-resource language (Arapaho), while for Tsez, our results are within 0.01 MER of GlossLM. Notably, CWoMP achieves these results using only IGT training data from the single lan-guage under consideration, while GlossLM benefits from pretraining on external multilingual glossed data (both methods incorporate multilingual foundation models—LaBSE and ByT5, respectively).

The only language where CWoMP underperforms is Uspanteko, which corresponds to a key qualitative observation—in the SIGMORPHON 2023 dataset, the Uspanteko data shows large variation in attested forms of morphemes. For example, the morpheme gloss *abuela* (Spanish for *grandmother*) is attested with segment *tiit'* in the train set and *tit'* in the test set; the latter is thus not correctly predicted. Frequent variations in morpheme forms are common in real-world IGT datasets, stemming from speaker variation and words that can be spelled in multiple ways, as well as speech errors and transcription errors. By design, CWoMP cannot generalize to unseen morphemes, which guarantees the outputs are free of hallucinated morphemes, but this means our method inherently cannot handle normal spelling variation unless it is attested in the training set. Thus, sensitivity to annotation consistency is an explicit limitation of CWoMP.

In all cases the extended lexicon setting outperforms the train lexicon setting, suggesting that many errors in the latter stem from morphemes that cannot be predicted because they are unattested in the training data. As such, the mean proportion of OOV morphemes per sentence should quantify the portion of MER that is structurally unrecoverable without lexicon expansion. We compare this OOV rate and MER gain from lexicon expansion in Table 2:

| | arp | usp | ddo | git | lez | ntu | nyb |
|---|---|---|---|---|---|---|---|
| $p_{\text{OOV}}$ | .043 | .069 | .007 | .446 | .064 | .042 | .021 |
| $\Delta_{\text{MER}}$ | .035 | .040 | .003 | .133 | .047 | .034 | .018 |

Table 2: Per-sentence proportion of OOV morphemes per language ($p_{\text{OOV}}$) and MER gain in the expanded lexicon setting ($\Delta_{\text{MER}}$). Language codes follow Table 1.

Overall, trends in these values broadly correspond with each other, the largest mismatch being for the lowest-resource language (Gitksan, with only 31 train samples) where prediction is harder overall due to limited training data.

We also provide qualitative examples (Appendix B.5) showing cases where baseline methods hallucinate morphemes unattested in training data, while CWoMP is constrained to select from its lexicon.

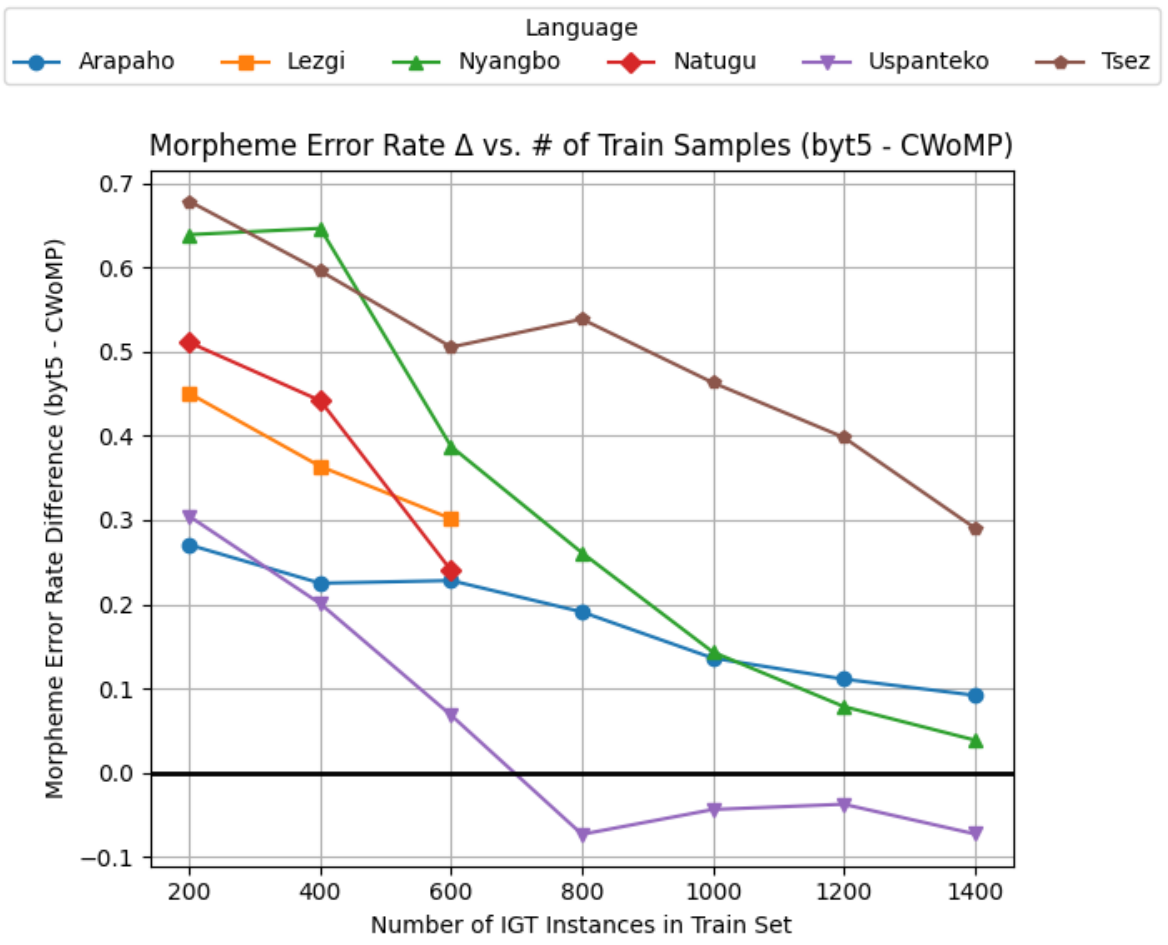

Figure 5: **CWoMP outperforms finetuned Byt5 in low-resource training scenarios.** Morpheme Error Rate Δ (ByT5 - CWoMP) for glossing as a function of number of training samples. Results are reported as the mean of 3 runs using independent random subsets.

## 4.4 Segmentation

**Task Description.** In addition to glossing, our model jointly predicts morphological segmentation from sentence transcripts and translations.

| | arp | usp | ddo | git | lez | ntu | nyb |
|---|---|---|---|---|---|---|---|
| ByT5 | <u>0.08</u> | **0.15** | 0.04 | 0.98 | 0.27 | **0.09** | <u>0.03</u> |
| CWoMP (Train) | 0.11 | 0.21 | <u>0.03</u> | <u>0.71</u> | <u>0.16</u> | 0.14 | 0.04 |
| CWoMP (Extended) | **0.07** | <u>0.17</u> | **0.02** | **0.57** | **0.11** | <u>0.10</u> | **0.02** |

Table 3: Morpheme Error Rate (↓) for **segmentation**. Best results are in **bold** and second-best <u>underlined</u>.

**Results.** We compare against a ByT5 (Xue et al., 2022) baseline finetuned on monolingual segmentation for each language in our test set. Table 3 shows that CWoMP demonstrates strong performance, particularly in the extended lexicon setting, outperforming the baseline for most languages.

A major advantage of CWoMP is its guaranteed alignment between morpheme segments and glosses, directly addressing the crucial usability issue identified by Rice et al. (2025) of misalignments in prior methods. Because CWoMP is constrained to retrieve attested segment/gloss pairs, it inherently cannot generate spurious pairings or mismatched numbers of segments and glosses.

## 4.5 Effect of Training Set Size

To evaluate how CWoMP scales with available data, we measure MER as a function of training set size. We compare against a finetuned ByT5 baseline, representative of the character-level sequence-to-sequence paradigm used in prior work (Ginn et al., 2024c). We use fixed hyperparameters across all data subsets to isolate the effect of training set size. Results are shown in Figure 5. Compared to the baseline, CWoMP performs noticeably better in the lowest-resource settings, with the gap narrowing as more data becomes available. This corresponds to the extremely resource-constrained language documentation setting where our method is most needed.

### 4.6 Computational Footprint

We compare the computational efficiency of our model against the previous SOTA—the ByT5-based GlossLM model. CWoMP has ∼1.2x fewer parameters (485M vs. 582M; counting both encoder and decoder) while requiring ∼8.5x fewer floating point operations (FLOPs) per sentence at inference time. This extreme gain in efficiency relative to the moderate reduction in parameters reflects our two-stage system. In standard autoregressive generation with GlossLM, the ByT5 decoder contributes to a majority (∼58%) of the total FLOPs in a typical generation, which scales with output length in bytes. On the other hand, CWoMP's small IGT decoder performs minimal computation, as each output is an entire morpheme; consequently, ∼88% of CWoMP's FLOPs are performed by the much larger BoM encoder—as the word being glossed must be embedded online before decoding. Overall, CWoMP performs less computation by avoiding ByT5's byte-level decoding bottleneck. Importantly, this comparison does not include the precomputation of our codebook of morpheme embeddings used for inference-time retrieval, which can be calculated and cached offline. For FLOP computation details, see Appendix A.4.

### 4.7 Embedding Interpretability

A key advantage of CWoMP over seq2seq baselines is that its predictions are grounded in a learned embedding space where words and morphemes have explicit geometric relationships. We examine whether these capture meaningful linguistic structure by testing for morphological analogies in the learned space, as shown in Figure 6.

As a concrete example, consider the Tsez word aždaħqay (segmentation `aždaħ-qay`, gloss `aždaħ-POSS.ABL`, translation `"from aždaħ's"`). Its nearest neighbors in the embedding space include other inflected forms of the same stem, like `aždaħqor` (similarity: 0.82) and `aždaħz` (similarity: 0.80), suggesting that morphological structure is represented geometrically. Moreover, the embedding space supports word2vec-style (Mikolov et al., 2013) morphological analogies in line with Ginn and Palmer (2024): the difference vectors are consistent across stems, such that *aždaħqay* − *aždaħqor* ≈ *babiwqay* − *babiwqor* ≈ *baruqay* − *baruqor*. This holds across all 20 stem pairs in the `POSS.ABL` to `POSS.LAT` transformation group, with a mean pairwise cosine similarity of 0.79 between difference vectors. This same structure holds for other transformation groups—three of which are shown in Figure 6, with mean cosine similarities ranging from 0.70 to 0.79.

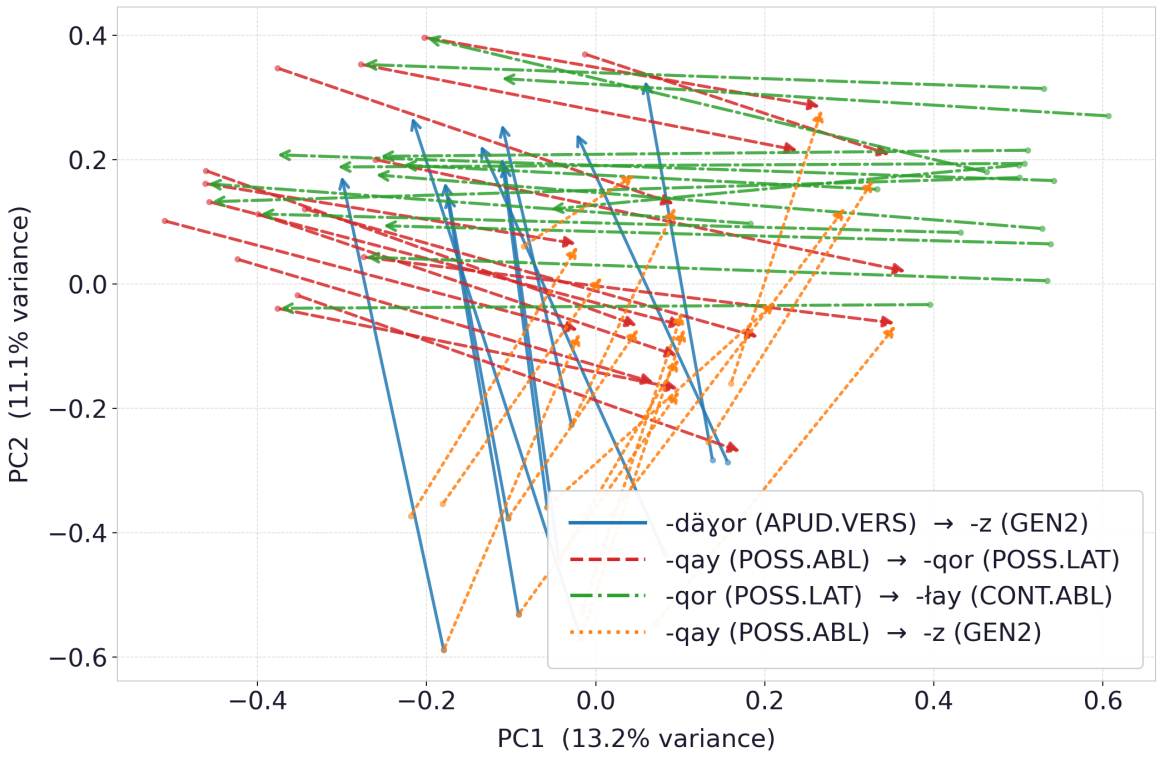

Figure 6: **Morphological analogies in Tsez word embedding space, visualised via PCA.** Points in the embedding space represent embeddings of Tsez words, calculated by CWoMP's BoM encoder. Each arrow connects two inflected forms of a common stem, pointing from the source to the target morphological form. The four colors represent four morphological transformations. Parallel arrows of the same color indicate that the difference vector between two inflected forms of a common stem is consistent across stems, suggesting linear morphological structure in the embedding space.

## 5 Related Work

**Automated Glossing.** Initial automated glossing systems used rule-based and statistical approaches (Palmer et al., 2009; Bender et al., 2014; Moeller and Hulden, 2018), while recent works have largely adopted neural methods, motivated by the SIGMORPHON 2023 shared task (Ginn et al., 2023), whose winning system (Girrbach, 2023) jointly learns to gloss and estimate surface segmentation using stochastic hard attention with straight-through gradient estimation. Most recent systems frame glossing as a sequence-to-sequence (seq2seq) task, predicting gloss strings learned via fine-tuning (Ginn et al., 2024c; He et al., 2024) or in-context learning (Ginn et al., 2024a,b). Unlike these systems that generate character or subword sequences, we leverage the morphological structure of the target language by learning and predicting sequences of morpheme-level representations. This prevents hallucination of unattested morpheme types and enables system updates at inference time without retraining. Separately from modeling contributions, Mortensen et al. (2023) address the annotation side of automated glossing, proposing an explicit, machine-readable convention (Generalized Glossing Guidelines) for representing non-concatenative morphological processes such as infixation and reduplication that are poorly handled by standard glossing formats; adapting CWoMP to such conventions is a promising direction for future work. Beyond modeling, Shandilya et al. (2026) build an annotation tool, GlossAssist, around CWoMP's retrieval-based predictions, closing the loop between automated glossing and the field linguists who ultimately use it.

**Representation Learning.** CWoMP draws on methods that align different views of data in shared embedding spaces via intra-modal similarity (Reimers and Gurevych, 2019; Feng et al., 2022) or cross-modal matching (Radford et al., 2021). While these typically use semantic similarity or align modalities such as text and vision, we adapt a similar dual encoder architecture and treat word in context and morphemes as parallel modalities.

Our learned morpheme embeddings also relate to work on distributional models of derivational morphology (Lazaridou et al., 2013; Kisselew et al., 2015), which showed that morphological composition can be modeled geometrically in embedding space. Recent work has shown that transformer embeddings capture compositional structure of compound words (Nagar et al., 2025). Most closely related to our approach, Chanchani and Huang (2023) maximize alignment between sentences and compositions of their phrasal constituents. Our approach differs from these prior works in learning morpheme representations via an explicit contrastive inclusion objective rather than analyzing emergent properties of pretrained embeddings.

**Generation in Learned Embedding Spaces.** Our autoregressive decoding of learned morpheme embeddings connects to two lines of work. Codebook-based generative models (Van Den Oord et al., 2017; Chen et al., 2025) learn discrete representations for autoregressive generation of continuous modalities such as images and audio. GIVT (Tschannen et al., 2024) extends this to continuous-valued outputs, more similar to the continuous outputs of our decoder. Generative retrieval models (Rajput et al., 2023; Li et al., 2025) generate token sequences that identify items to retrieve. Among these, SimCIT (Zhai et al., 2025) is most similar to us in using contrastively-learned cross-modal representations as tokens. Our setting differs from both of these directions: we operate on linguistically meaningful discrete units, each decoder output independently selects a morpheme rather than contributing to a composite identifier; and the lexicon can be expanded at inference time, unlike the fixed codebooks of prior approaches.

We also differ from retrieval-augmented generation (RAG) approaches (Lewis et al., 2020) that retrieve items with learned embeddings to condition a downstream generator. By contrast, CWoMP integrates retrieval directly into the generation mechanism, selecting an entry from the morpheme lexicon at each step. This is conceptually related to pointer networks (Vinyals et al., 2015) which use output embeddings to select from a dynamic set. However, our method selects from an external, learned mutable lexicon rather than from the input sequence.

## 6 Conclusion

We present CWoMP, a retrieval-based system for morphological glossing and segmentation. Our results demonstrate strong performance over existing gloss generation models, while providing several user-centric benefits: outputs are free of hallucinated morpheme types, segmentation and glossing are inherently aligned, and the morpheme lexicon is extensible at inference time without retraining.

In the future, we will investigate how CWoMP might support real users through human-in-the-loop documentary linguistic workflows. Other directions include incorporating additional linguistic resources such as dictionaries, formal grammars, and cross-lingual data (Tanzer et al., 2024; Zhang et al., 2024; Shandilya and Palmer, 2025). More broadly, we also envision that CWoMP might be applicable to structured prediction tasks beyond gloss generation such as morphological feature tagging in Universal Dependencies, or the annotation of biological sequences.

## Limitations

Our method has several inherent limitations. We assume that IGT is formatted in two corresponding tiers (segmentation and gloss) with equal numbers of units separated by hyphens. While prior methods learn from glosses alone, CWoMP also uses segmentation as supervision; while this enables leveraging this additional signal for training, it also limits the method's applicability to settings where underlying segmentation is available. CWoMP may be sensitive to inconsistent morpheme annotations or use of surface segmentation with substantial variation in morpheme forms, as discussed in Section 4.3. We have also not yet included conventions such as clitics separated by the = symbol, which future work could address. While CWoMP is more constrained and interpretable than methods that generate glosses as free text, it is also limited in being unable to generate forms of new morphemes that are not present in the lexicon (unless it is extended with these forms). While we test performance on diverse languages with different typologies, further work could explore its performance on languages with additional typological features that are under-represented in the SIGMORPHON 2023 dataset, such as extensive non-concatenative morphology. Finally, as with other generative applications of machine learning, care must be taken to validate outputs of our system with expert speakers and linguists, to avoid propagating incorrect language materials for endangered languages.

## Acknowledgments

We thank Noam Ordan and Lindia Tjuatja for providing helpful feedback. This work was supported by the National Science Foundation under Grants No. 2149404, "CAREER: From One Language to Another." and No. 2211951, "From Acoustic Signal to Morphosyntactic Analysis in One End-to-End Neural System." Any opinions, findings, and conclusions or recommendations expressed in this material are those of the authors and do not necessarily reflect the views of the National Science Foundation. Parts of this work were supported by the Blanca condo computing resource at the University of Colorado Boulder. Blanca is jointly funded by computing users and the University of Colorado Boulder.

## References

A. K. Abdulaev and I. K. Abdullaev, editors. 2010. *Cezyas folklor / Dido (Tsez) folklore / Didojskij (cezskij) fol'klor*. Lotos, Leipzig–Makhachkala.

Khuyagbaatar Batsuren, Omer Goldman, Salam Khalifa, Nizar Habash, Witold Kieraś, Gábor Bella, Brian Leonard, Garrett Nicolai, Kyle Gorman, Yustinus Ghanggo Ate, Maria Ryskina, Sabrina Mielke, Elena Budianskaya, Charbel El-Khaissi, Tiago Pimentel, Michael Gasser, William Abbott Lane, Mohit Raj, Matt Coler, and 76 others. 2022. UniMorph 4.0: Universal Morphology. In *Proceedings of the Thirteenth Language Resources and Evaluation Conference*, pages 840–855, Marseille, France. European Language Resources Association.

Emily M Bender, Joshua Crowgey, Michael Wayne Goodman, and Fei Xia. 2014. Learning grammar specifications from igt: A case study of chintang. In *Proceedings of the 2014 Workshop on the Use of Computational Methods in the Study of Endangered Languages*, pages 43–53.

Sachin Chanchani and Ruihong Huang. 2023. Composition-contrastive learning for sentence embeddings. In *Proceedings of the 61st Annual Meeting of the Association for Computational Linguistics (Volume 1: Long Papers)*, pages 15836–15848.

Sanyuan Chen, Chengyi Wang, Yu Wu, Ziqiang Zhang, Long Zhou, Shujie Liu, Zhuo Chen, Yanqing Liu, Huaming Wang, Jinyu Li, Lei He, Sheng Zhao, and Furu Wei. 2025. Neural codec language models are zero-shot text to speech synthesizers. *IEEE Transactions on Audio, Speech and Language Processing*, 33:705–718.

Jonathan H Clark, Dan Garrette, Iulia Turc, and John Wieting. 2022. Canine: Pre-training an efficient tokenization-free encoder for language representation. *Transactions of the Association for Computational Linguistics*, 10:73–91.

Marie-Catherine de Marneffe, Christopher D. Manning, Joakim Nivre, and Daniel Zeman. 2021. Universal Dependencies. *Computational Linguistics*, 47(2):255–308.

Fangxiaoyu Feng, Yinfei Yang, Daniel Cer, Naveen Arivazhagan, and Wei Wang. 2022. Language-agnostic BERT sentence embedding. In *Proceedings of the 60th Annual Meeting of the Association for Computational Linguistics (Volume 1: Long Papers)*, pages 878–891, Dublin, Ireland. Association for Computational Linguistics.

Luke Gessler. 2022. Closing the NLP gap: Documentary linguistics and NLP need a shared software infrastructure. In *Proceedings of the Fifth Workshop on the Use of Computational Methods in the Study of Endangered Languages*, pages 119–126, Dublin, Ireland. Association for Computational Linguistics.

Michael Ginn. 2023. Sigmorphon 2023 shared task of interlinear glossing: Baseline model. *Preprint*, arXiv:2303.14234.

Michael Ginn, Mans Hulden, and Alexis Palmer. 2024a. Can we teach language models to gloss endangered languages? In *Findings of the Association for Computational Linguistics: EMNLP 2024*, pages 5861–5876.

Michael Ginn, Ali Marashian, Bhargav Shandilya, Claire Post, Enora Rice, Juan Vásquez, Marie Mcgregor, Matthew Buchholz, Mans Hulden, and Alexis Palmer. 2024b. On the robustness of neural models for full sentence transformation. In *Proceedings of the 4th Workshop on Natural Language Processing for Indigenous Languages of the Americas (AmericasNLP 2024)*, pages 159–173.

Michael Ginn, Sarah Moeller, Alexis Palmer, Anna Stacey, Garrett Nicolai, Mans Hulden, and Miikka Silfverberg. 2023. Findings of the SIGMORPHON 2023 shared task on interlinear glossing. In *Proceedings of the 20th SIGMORPHON workshop on Computational Research in Phonetics, Phonology, and Morphology*, pages 186–201, Toronto, Canada. Association for Computational Linguistics.

Michael Ginn and Alexis Palmer. 2024. Decomposing fusional morphemes with vector embeddings. In *Proceedings of the 21st SIGMORPHON workshop on Computational Research in Phonetics, Phonology, and Morphology*, pages 57–66, Mexico City, Mexico. Association for Computational Linguistics.

Michael Ginn, Lindia Tjuatja, Taiqi He, Enora Rice, Graham Neubig, Alexis Palmer, and Lori Levin. 2024c. GlossLM: A massively multilingual corpus and pretrained model for interlinear glossed text. In *Proceedings of the 2024 Conference on Empirical Methods in Natural Language Processing*, pages 12267–12286, Miami, Florida, USA. Association for Computational Linguistics.

Michael Ginn, Lindia Tjuatja, Enora Rice, Ali Marashian, Maria Valentini, Jasmine Xu, Graham Neubig, and Alexis Palmer. 2026. Massively multilingual joint segmentation and glossing. In *Proceedings of the 64th Annual Meeting of the Association for Computational Linguistics (Volume 1: Long Papers)*, pages 35806–35820, San Diego, California, United States. Association for Computational Linguistics.

Leander Girrbach. 2023. Tü-cl at sigmorphon 2023: Straight-through gradient estimation for hard attention. In *Proceedings of the 20th SIGMORPHON workshop on Computational Research in Phonetics, Phonology, and Morphology*, pages 151–165.

Taiqi He, Kwanghee Choi, Lindia Tjuatja, Nathaniel Robinson, Jiatong Shi, Shinji Watanabe, Graham Neubig, David Mortensen, and Lori Levin. 2024. Wav2Gloss: Generating interlinear glossed text from speech. In *Proceedings of the 62nd Annual Meeting of the Association for Computational Linguistics (Volume 1: Long Papers)*, pages 568–582, Bangkok, Thailand. Association for Computational Linguistics.

Prannay Khosla, Piotr Teterwak, Chen Wang, Aaron Sarna, Yonglong Tian, Phillip Isola, Aaron Maschinot, Ce Liu, and Dilip Krishnan. 2020. Supervised contrastive learning. *Advances in neural information processing systems*, 33:18661–18673.

Diederik P. Kingma and Jimmy Ba. 2014. Adam: A method for stochastic optimization. *CoRR*, abs/1412.6980.

Max Kisselew, Sebastian Padó, Alexis Palmer, and Jan Šnajder. 2015. Obtaining a better understanding of distributional models of german derivational morphology. In *Proceedings of the 11th international conference on computational semantics*, pages 58–63.

Angeliki Lazaridou, Marco Marelli, Roberto Zamparelli, and Marco Baroni. 2013. Compositional-ly derived representations of morphologically complex words in distributional semantics. In *Proceedings of the 51st annual meeting of the association for computational linguistics (volume 1: Long papers)*, pages 1517–1526.

Christian Lehmann. 1982. Directions for interlinear morphemic translations. 16(1–4):199–224.

Patrick Lewis, Ethan Perez, Aleksandra Piktus, Fabio Petroni, Vladimir Karpukhin, Naman Goyal, Heinrich Küttler, Mike Lewis, Wen-tau Yih, Tim Rocktäschel, and 1 others. 2020. Retrieval-augmented generation for knowledge-intensive nlp tasks. *Advances in neural information processing systems*, 33:9459–9474.

Xiaoxi Li, Jiajie Jin, Yujia Zhou, Yuyao Zhang, Peitian Zhang, Yutao Zhu, and Zhicheng Dou. 2025. From matching to generation: A survey on generative information retrieval. *ACM Transactions on Information Systems*, 43(3):1–62.

Manuel Mager, Özlem Çetinoğlu, and Katharina Kann. 2020. Tackling the low-resource challenge for canonical segmentation. In *Proceedings of the 2020 Conference on Empirical Methods in Natural Language Processing (EMNLP)*, pages 5237–5250, Online. Association for Computational Linguistics.

Tomas Mikolov, Kai Chen, Greg Corrado, and Jeffrey Dean. 2013. Efficient estimation of word representations in vector space. *Preprint*, arXiv:1301.3781.

Sarah Moeller and Mans Hulden. 2018. Automatic glossing in a low-resource setting for language documentation. In *Proceedings of the Workshop on Computational Modeling of Polysynthetic Languages*, pages 84–93.

David R. Mortensen, Ela Gulsen, Taiqi He, Nathaniel Robinson, Jonathan Amith, Lindia Tjuatja, and Lori Levin. 2023. Generalized glossing guidelines: An explicit, human- and machine-readable, item-and-process convention for morphological annotation. In *Proceedings of the 20th SIGMORPHON workshop on Computational Research in Phonetics, Phonology, and Morphology*, pages 58–67, Toronto, Canada. Association for Computational Linguistics.

Aishik Nagar, Ishaan Singh Rawal, Mansi Dhanania, and Cheston Tan. 2025. How do transformer embeddings represent compositions? a functional analysis. In *Findings of the Association for Computational Linguistics: ACL 2025*, pages 21444–21461.

Aaron van den Oord, Yazhe Li, and Oriol Vinyals. 2018. Representation learning with contrastive predictive coding. *arXiv preprint arXiv:1807.03748*.

Alexis Palmer, Taesun Moon, and Jason Baldridge. 2009. Evaluating automation strategies in language documentation. In *Proceedings of the NAACL HLT 2009 Workshop on Active Learning for Natural Language Processing*, pages 36–44.

Aidan Pine and Mark Turin. 2017. Language revitalization. *Oxford research encyclopedia of linguistics*.

Alec Radford, Jong Wook Kim, Chris Hallacy, Aditya Ramesh, Gabriel Goh, Sandhini Agarwal, Girish Sastry, Amanda Askell, Pamela Mishkin, Jack Clark, and 1 others. 2021. Learning transferable visual models from natural language supervision. In *International conference on machine learning*, pages 8748–8763. PmLR.

Shashank Rajput, Nikhil Mehta, Anima Singh, Raghunandan Hulikal Keshavan, Trung Vu, Lukasz Heldt, Lichan Hong, Yi Tay, Vinh Tran, Jonah Samost, and 1 others. 2023. Recommender systems with generative retrieval. *Advances in Neural Information Processing Systems*, 36:10299–10315.

Nils Reimers and Iryna Gurevych. 2019. Sentence-bert: Sentence embeddings using siamese bert-networks. In *Proceedings of the 2019 conference on empirical methods in natural language processing and the 9th international joint conference on natural language processing (EMNLP-IJCNLP)*, pages 3982–3992.

Enora Rice, Katharina von der Wense, and Alexis Palmer. 2025. Interdisciplinary research in conversation: A case study in computational morphology for language documentation. In *Proceedings of the 2025 Conference on Empirical Methods in Natural Language Processing*, pages 11273–11285, Suzhou, China. Association for Computational Linguistics.

Frank Seifart, Nicholas Evans, Harald Hammarström, and Stephen C Levinson. 2018. Language documentation twenty-five years on. *Language*, 94(4):e324–e345.

Bhargav Shandilya, Matt Buchholz, and Alexis Palmer. 2026. Glossassist: A tool to simplify corpus creation and study the effect of nlp models in low-resource documentation settings. *Preprint*, arXiv:2606.04367.

Bhargav Shandilya and Alexis Palmer. 2025. Boosting the capabilities of compact models in low-data contexts with large language models and retrieval-augmented generation. In *Proceedings of the 31st International Conference on Computational Linguistics*, pages 7470–7483, Abu Dhabi, UAE. Association for Computational Linguistics.

Garrett Tanzer, Mirac Suzgun, Eline Visser, Dan Jurafsky, and Luke Melas-Kyriazi. 2024. A benchmark for learning to translate a new language from one grammar book. In *The Twelfth International Conference on Learning Representations*.

Michael Tschannen, Cian Eastwood, and Fabian Mentzer. 2024. Givt: Generative infinite-vocabulary transformers. In *European Conference on Computer Vision*, pages 292–309. Springer.

Aaron Van Den Oord, Oriol Vinyals, and 1 others. 2017. Neural discrete representation learning. *Advances in neural information processing systems*, 30.

Oriol Vinyals, Meire Fortunato, and Navdeep Jaitly. 2015. Pointer networks. *Advances in neural information processing systems*, 28.

Linting Xue, Aditya Barua, Noah Constant, Rami Al-Rfou, Sharan Narang, Mihir Kale, Adam Roberts, and Colin Raffel. 2022. ByT5: Towards a token-free future with pre-trained byte-to-byte models. *Transactions of the Association for Computational Linguistics*, 10:291–306.

Penglong Zhai, Yifang Yuan, Fanyi Di, Jie Li, Yue Liu, Chen Li, Jie Huang, Sicong Wang, Yao Xu, and Xin Li. 2025. A simple contrastive framework of item tokenization for generative recommendation. *arXiv preprint arXiv:2506.16683*.

Kexun Zhang, Yee Man Choi, Zhenqiao Song, Taiqi He, William Yang Wang, and Lei Li. 2024. Hire a linguist!: Learning endangered languages in LLMs with in-context linguistic descriptions. In *Findings of the Association for Computational Linguistics: ACL 2024*, pages 15654–15669, Bangkok, Thailand. Association for Computational Linguistics.

Xingyuan Zhao, Satoru Ozaki, Antonios Anastasopoulos, Graham Neubig, and Lori Levin. 2020. Automatic interlinear glossing for under-resourced languages leveraging translations. In *Proceedings of the 28th International Conference on Computational Linguistics*, pages 5397–5408, Barcelona, Spain (Online). International Committee on Computational Linguistics.

# A Additional Implementation Details

## A.1 Training Hyperparameters

Below we provide hyperparameter values, noting defaults and language-specific deviations. The latter are determined by coarse hyperparameter searches on validation data. After training for all epochs, we use checkpoints from the best epoch according to validation metrics for final testing.

**BoM Encoder** We train our final encoder with a batch size of 128 for 100 epochs (1000 for Gitksan) with a learning rate of $2 \times 10^{-5}$. We employ a learning rate scheduler with 100 warmup steps.

**IGT Decoder** Our decoder is initialized 4 decoder blocks with a hidden dimension of 512 and 4 self-attention heads. We train with a batch size of 32 for 100 epochs (1000 for Gitksan) using a learning rate of $1 \times 10^{-4}$. Our decoder is optimized using a weight decay of 0.01 and gradient clipping (max norm 1.0). We set dropout to 0.1.

**ByT5 Baseline** We train ByT5 for a max 30 epochs using early stopping with patience 5. We use a batch size of 64 and a learning rate of $5 \times 10^{-5}$. We use Adam optimization (Kingma and Ba, 2014) and gradient clipping (max norm 1.0). We use the same hyperparameters for both segmentation and glossing.

### A.2 Inference Settings

For IGT decoding, we use beam search with 5 beams.

### A.3 Metric Calculations

MER follows the implementation of Ginn et al. (2026), including clipping with maximum value 1 and excluding punctuation-only tokens from calculations. We aggregate MER on the sample level (not corpus level).

### A.4 FLOPs calculation

We estimate FLOPs per sentence using the standard transformer multiply-accumulate formula covering self-attention and feed-forward layers. Embeddings, layer norms, and output projections are excluded as minor contributors. For CWoMP, total FLOPs per sentence combine the BoM encoder, which encodes all words in the sentence in a single batched forward pass, and the IGT decoder, which performs beam search autoregressively for each word. For GlossLM, total FLOPs combine the ByT5 encoder, which runs once per sentence and is shared across all beams, and the ByT5 decoder, which runs beam search over the full byte-level output sequence. The decoder FLOPs account for both self-attention and cross-attention at each step. Both models are evaluated under their actual inference beam sizes (GlossLM: 3, CWoMP: 5), and sentence-level estimates assume 8 words per sentence, 3 morphemes per word, and 128 decoder output bytes for GlossLM. Under these assumptions, CWoMP requires 24.8B FLOPs per sentence (BoM encoder: 21.8B, IGT decoder: 3.0B) compared to 211.7B for GlossLM (ByT5 encoder: 87.8B, ByT5 decoder: 123.9B).

## B Additional Results

### B.1 Morpheme Retrieval

**Task Description.** This task directly tests the performance of CWoMP's BoM encoder module (Section 3.2). We evaluate its final test performance on *word→morpheme* retrieval, reporting standard retrieval metrics: recall@k, Normalized Distributed Cumulative Gain (NDCG) and Mean Average Precision (mAP). The target set to retrieve consists of all ground-truth morphemes in the word (restricted to in-vocabulary morphemes in the train lexicon setting). This task serves two purposes—it validates the encoder's performance before training the downstream decoder, and may be useful in its own right for future morphological tasks where analysis should be restricted to retrieved morphemes.

**Results.** Retrieval results are shown in Table 4. The BoM encoder generally shows strong performance at retrieving relevant morphemes, and further benefits from the extended lexicon setting.

| | arp | usp | ddo | git | lez | ntu | nyb |
|---|---|---|---|---|---|---|---|
| *Train Lexicon* | | | | | | | |
| P@1 | .829 | .644 | .901 | .624 | .880 | .930 | .907 |
| R@10 | .950 | .920 | .984 | .923 | .965 | .990 | .990 |
| NDCG | .874 | .784 | .934 | .781 | .917 | .959 | .945 |
| mAP | .823 | .721 | .899 | .717 | .886 | .938 | .918 |
| *Extended Lexicon* | | | | | | | |
| P@1 | .859 | .656 | .901 | .732 | .896 | .929 | .910 |
| R@10 | .952 | .927 | .984 | .899 | .970 | .989 | .991 |
| NDCG | .885 | .796 | .934 | .810 | .925 | .959 | .947 |
| mAP | .833 | .733 | .899 | .753 | .893 | .937 | .921 |

Table 4: Retrieval performance of our final encoder configuration. NDCG and R are @10; mAP is @100. Language codes are as in Table 1.

### B.2 Additional Metrics

In addition to the results in our main paper using a MER metric, here we report additional metrics following prior work: For glossing, Table 5 reports morpheme and word accuracy metrics from Ginn et al. (2024c); for segmentation, Table 6 uses the modified F1 score and whole-word accuracy of Mager et al. (2020). Glossing metrics are reported at the corpus level. Segmentation metrics are computed at the sample level and averaged. Overall, the relative performance of models is consistent with our main results, justifying our use of MER (see Section 4.1 for our conceptual motivation for this choice of metric).

| | arp | usp | ddo | git | lez | ntu | nyb |
|---|---|---|---|---|---|---|---|
| *Morpheme Accuracy (%)* ↑ | | | | | | | |
| GlossLM (+ft) | 82.1 | **78.6** | **83.6** | 10.1 | 57.3 | 62.8 | 87.4 |
| CWoMP (Train) | 80.6 | 69.8 | 82.8 | 11.1 | 66.0 | 66.6 | 88.2 |
| CWoMP (Ext.) | **84.1** | 72.9 | 82.9 | **14.8** | **69.3** | **67.8** | **89.1** |
| *Word Accuracy (%)* ↑ | | | | | | | |
| GlossLM (+ft) | 81.5 | **81.0** | 87.3 | 28.4 | 64.9 | 78.9 | 86.2 |
| CWoMP (Train) | 83.7 | 75.3 | 86.9 | 24.7 | 77.3 | 82.4 | 86.8 |
| CWoMP (Ext.) | **87.9** | 79.9 | **87.4** | **38.3** | **81.9** | **86.1** | **87.8** |

Table 5: Morpheme and Word Accuracy (↑) on Glossing. Best results are in **bold** and second-best underlined. Language codes follow Table 1.

| | arp | usp | ddo | git | lez | ntu | nyb |
|---|---|---|---|---|---|---|---|
| *F1* ↑ | | | | | | | |
| Finetuned ByT5 | .939 | **.872** | .977 | .098 | .756 | .923 | .976 |
| CWoMP (Train) | .908 | .807 | .977 | .377 | .873 | .900 | .969 |
| CWoMP (Ext.) | **.943** | .850 | **.981** | **.533** | **.917** | **.931** | **.983** |
| *Accuracy* ↑ | | | | | | | |
| Finetuned ByT5 | .768 | .551 | .690 | .000 | .149 | .374 | .920 |
| CWoMP (Train) | .860 | .768 | .957 | .219 | .831 | .866 | .953 |
| CWoMP (Ext.) | **.909** | **.811** | **.963** | **.407** | **.880** | **.904** | **.968** |

Table 6: F1 and Accuracy (↑) on Segmentation. Best results are in **bold** and second-best underlined. Language codes follow Table 1.

### B.3 Additional Comparison

We provide a comparison to the concurrent Polygloss model of Ginn et al. (2026) in Table 7 (using their most performant model—ByT5 Interleaved). We reproduce their reported MER scores for our languages, along with our performance. For a fair comparison, we use Polygloss' training and evaluation data, which does not exactly match the SIGMORPHON 2023 shared task data. **Therefore, the metrics reported in Table 7 are not directly comparable to the metrics reported in Figure 4.**

| | arp | usp | ddo | git | lez | ntu | nyb |
|---|---|---|---|---|---|---|---|
| Polygloss | .152 | **.160** | **.072** | .597 | .357 | .142 | .222 |
| CWoMP (Train) | .141 | .215 | .086 | .670 | .345 | .161 | .220 |
| CWoMP (Extended) | **.094** | .177 | .082 | **.523** | **.259** | **.126** | **.186** |

Table 7: Comparison between Polygloss (Ginn et al., 2026) and CWoMP, reporting Morpheme Error Rate (↓) for Glossing. Best results are in **bold** and second-best underlined. Language codes follow Table 1.

We find that CWoMP exceeds Polygloss on several languages, particularly in the extended lexicon setting.

### B.4 Ablations

We ablate the following key design choices of our method:

- **Base encoder:** CANINE-S (Clark et al., 2022) vs. LaBSE (Feng et al. 2022; used in our full system)
- **Loss function:** Standard InfoNCE vs. our multi-positive variant (Section 3.2)
- **Prompt format:** Inclusion of translations and transcripts in prompts for encoding words; use of forced character tokenization via spaces.

For these ablations, encoder models are trained for 50 epochs with a batch size of 64. All other hyperparameters are equivalent to those used for our final configuration (Appendix A.1). Ablations are run for Tsez, Gitksan, Natugu and Nyangbo.

We report validation metrics in Table 8, seeing that each component of our system has an overall positive impact on results, with our full system mostly outperforming alternative configurations.

| Encoder | Loss | Prompt | NDCG | mAP | P@1 | R@10 |
|---|---|---|---|---|---|---|
| CANINE-S | InfoNCE | W | .837 | .774 | .761 | .894 |
| | | W+Tr | .838 | .777 | .766 | .933 |
| | | W+Ts | .827 | .765 | .758 | .931 |
| | | W+Tr+Ts | .847 | .791 | .776 | .937 |
| LaBSE | InfoNCE | W | .832 | .792 | .773 | .908 |
| | | W+Tr | .846 | .805 | .795 | .916 |
| | | W+Ts | .866 | .823 | .816 | .938 |
| | | W+Tr+Ts | .881 | .844 | **.840** | .942 |
| | | W+Tr+Ts+Sp | .890 | .846 | .821 | .970 |
| | MultPos | W+Tr+Ts+Sp | **.897** | **.856** | .826 | **.972** |

Table 8: Encoder, prompt format and loss function ablations. Metrics (↑) averaged across languages; NDCG is @10 and mAP is @100. Prompt format: W=Word, Tr=Transcript, Ts=Translation, Sp=character spacing. Best results are in **bold** and second-best underlined.

### B.5 Sample Predictions

We provide representative system outputs in Table 12, and highlight specific qualitative patterns below.

Beyond the qualitative patterns discussed below, hallucination is also substantial in quantitative terms. Across our languages, GlossLM's predicted glosses include an illegal gloss (i.e., one absent from the training vocabulary). The morpheme-level hallucination rates are 4.48% for Arapaho, 44.84% for Gitksan, 7.07% for Lezgi, 5.31% for Natugu, 0.42% for Nyangbo, 0.95% for Tsez, and 4.02% for Uspanteko. Overall, hallucination is not very common in six of the seven languages, with a micro-average rate of 4.27% across all predictions. Gitksan is the exception, likely because its training set has only very few sentences and provides limited vocabulary coverage. These results suggest that GlossLM generally relies on morphemes observed in the language-specific training data. Hallucination becomes a substantial problem in extremely low-resource settings.

Tables 9 to 11 illustrate three notable hallucination patterns in GlossLM outputs. Table 9 shows GlossLM hallucinating a morpheme that was neither observed during training nor appropriate in context. Table 10 shows GlossLM generating grammatical tags that deviate from the expected convention (e.g., PST.UW instead of PST.UNW). Because CWoMP is constrained to a discrete codebook, its outputs exactly match attested or user-provided glosses, precluding such errors. We note, however, that GlossLM can sometimes correctly predict glosses unseen during training, such as "rough" in Table 10 and "police" in Table 11—likely due to sentential context (e.g., the translation tier) or string matching ("police" is an English loanword in Gitksan). In both cases, CWoMP fails to produce the correct label—a limitation that can be mitigated by inference-time lexicon expansion.

## C Generative AI Disclosure

Generative AI tools were used for polishing wording and formatting in this manuscript, and for coding assistance.

| | |
|---|---|
| Transcript | Kaƛ'iƛin , žawab teƛno nesä . |
| Segmentation | koƛ'i-ƛin žawab teƛ-n nesi-a |
| Translation | "I can.", he answered. |
| Gold Gloss | understand.to-QUOT answer give-PST.UNW DEM1.ISG.OBL-ERG |
| GlossLM Gloss | unable-QUOT answer give-PST.UNW DEM1.ISG.OBL-ERG |
| CWoMP (Train) Gloss | understand.to-QUOT answer give-PST.UNW DEM1.ISG.OBL-ERG |
| Unattested Labels in Gold | N/A |
| Unattested GlossLM Labels | unable |

Table 9: **GlossLM can mistakenly hallucinate gloss labels not seen during training.** In this Tsez example, GlossLM outputs the label "unable" despite being unattested during training and contextually inappropriate. Incorrect predictions are highlighted.

| | |
|---|---|
| Transcript | Xediw qač'aw , qizanyos t'alam bodixanusi žek'u zown |
| Segmentation | xediw qač'aw qizan-s t'alam b-odi-xanusi žek'u zow-n |
| Translation | The husband was a rough man who did not do the family's duty. |
| Gold Gloss | husband rough family-GEN1 care III-do-NEG.PRS.PRT man be.NPRS-PST.UNW |
| GlossLM Gloss | husband rough family-GEN1 duty III-do- NEG.PRS.PTCP man be.NPRS- PST.UW |
| CWoMP (Train) Gloss | husband living family-GEN1 care III-do-NEG.PRS.PRT man be.NPRS-PST.UNW |
| CWoMP (Extended) Gloss | husband rough family-GEN1 care III-do-NEG.PRS.PRT man be.NPRS-PST.UNW |
| Unattested Labels in Gold | rough |
| Unattested GlossLM Labels | NEG.PRS.PTCP, PST.UW |

Table 10: **GlossLM can mistakenly generate grammatical tags deviating from expected conventions.** In this Tsez example, GlossLM generates PST.UW instead of PST.UNW and NEG.PRS.PTCP instead of NEG.PRS.PRT. Incorrect predictions are highlighted.

| | |
|---|---|
| Transcript | Needii 'nakwt ii bakwhl police. |
| Segmentation | nee-dii 'nekw-t ii bakw-hl police |
| Translation | Not long after, the police came. |
| Gold Gloss | NEG-FOC long-3.II CCNJ come.PL-CN police |
| GlossLM Gloss | NEG-FOC long.after CCNJ arrive-CN police |
| CWoMP (Train) Gloss | NEG-FOC long-long CCNJ arrive.PL-CN pole |
| CWoMP (Extended) Gloss | NEG-FOC long-long CCNJ come.PL-CN police |
| Unattested Labels in Gold | come.PL, police |
| Unattested GlossLM Labels | long.after, police |

Table 11: **GlossLM can correctly predict unattested labels.** In this Gitksan example, GlossLM predicts "police" despite it not being present attested in the train lexicon. Note that "police" is an English loanword in Gitksan, and further appears in the translation tier. Incorrect predictions are highlighted.

Table 12: Random selection of system outputs (Part 1 of 2)

| | |
|---|---|
| Language | Natugu |
| Transcript | Nzpwrkilvcngr doa kcng nztupebz badr |
| Seg. | nz-pwrkilvc-ngr doa kcng nz-tu-pe-bz badr |
| Translation | Those students were surprised about him standing there with them |
| Gold Gloss | 3AUG-startled-APPL person those PAS-stand-COS-PDIR.YON COM.PL |
| GlossLM Gloss | 3AUG-be.surprised-about-APPL child those 3AUG-stand-COS-PDIR.YON COM.P |
| CWoMP (Train) Seg. | nzvecngr-ki-ngr doa kcng nz-tu-pe-bz badr |
| CWoMP (Train) Gloss | war-path-APPL child those 3AUG-stand-COS-PDIR.YON COM.PL' |
| CWoMP (Ext.) Seg. | nzvecngr-ki-ngr doa kcng nz-tu-pe-bz badr |
| CWoMP (Ext.) Gloss | war-path-APPL child those 3AUG-stand-COS-PDIR.YON COM.PL |
| Language | Natugu |
| Transcript | Rpi-mopwz kxetu r ami kx , " Nim ncblo kznike ?" |
| Seg. | rpi-mou-bz kxetu r ami kx , " nim ncblo kznike ?" |
| Translation | The head of the army further said, "What kind of man are you?" |
| Gold Gloss | said-again-PDIR.YON leader GEN1A army SUBR , " be.2MINII mankind who ?" |
| GlossLM Gloss | said-again-PDIR.YON leader GEN1A army SUBR be.2MINII mankind kind? |
| CWoMP (Train) Seg. | rpi-mou-bz kx-etu r ami kx , " nim ncblo nike ?" |
| CWoMP (Train) Gloss | said-again-PDIR.YON SUBR-be.big GEN1A army SUBR , " be.2MINII mankind what ?" |
| CWoMP (Ext.) Seg. | rpi-mou-bz kx-etu r ami kx , " nim ncblo kznike ?" |
| CWoMP (Ext.) Gloss | said-again-PDIR.YON SUBR-be.big GEN1A army SUBR , " be.2MINII mankind who ?" |
| Language | Gitksan |
| Transcript | "Ii hasag̲a'm dim hogwin litxwhl k̲'ay limxs̲im gyet dim ent hlimoo'm." |
| Seg. | ii hasak̲-'m dim hogwin lit-xw-hl k̲'ay limxs̲-m gyet dim en-t hlimoo-'m |
| Translation | "We will want the support of young men to help us." |
| Gold Gloss | CCNJ want-1PL.II PROSP near stand-PASS-CN still grow-ATTR man PROSP AX-3.I help-1PL.II |
| GlossLM Gloss | CCNJ want-1PL.II PROSP house-1PL.II PROSP-CN young.man PROSP PROSP help-1PL.II |
| CWoMP (Train) Seg. | ii baasax̲ dim g̲oo-xwin-xwin hahla'lst-hl k̲'ak̲ lax̲'ni-'m guu-diit dim nee-t hahla'lst-'m-'m |
| CWoMP (Train) Gloss | CCNJ separate PROSP LOC-DEM.PROX-DEM.PROX work-CN open hear-1PL.II take-3PL.II PROSP NEG-DM work-1PL.II-1PL.II |
| CWoMP (Ext.) Seg. | ii hasak̲ dim hogwin-hogwin-xwin luuhligyootxw-hl ky'ax̲ limx̲s-'m gu-t dim nee-t hlimoo-'m-'m |
| CWoMP (Ext.) Gloss | CCNJ desire PROSP toward-toward-DEM.PROX axe-CN unanimously grow-1PL.II what-3PL.II PROSP NEG-DM help-1PL.II-1PL.II |
| Language | Gitksan |
| Transcript | Ii sit'aa'mam depdiithl hlidaaxhl hlgu lax̲ha'niijok̲. |
| Seg. | ii si-t'aa-'ma-m dep-diit-hl hlidaax-hl hlgu lax̲ha-'nii-jok̲ |
| Translation | They started measuring out the little settlement. |
| Gold Gloss | CCNJ CAUS1-sit-1PL.II-ATTR measure-3PL.II-CN circumference-CN little on INS-on-live |
| GlossLM Gloss | CCNJ start-PROG.3PL measure-CN house-CN DEM DIM VIS.INFR-forget.PL |
| CWoMP (Train) Seg. | ii 'wa sdil-diit hahla'lst-hl hlgu lax̲ g̲a-'nakw |
| CWoMP (Train) Gloss | CCNJ reach accompany-3PL.II work-CN small on DISTR-long |
| CWoMP (Ext.) Seg. | ii 'ma diye-diit hlidaax-hl hlgu lax̲ g̲a-'nakw |
| CWoMP (Ext.) Gloss | CCNJ DETR QUOT.3SG-3PL.II circumference-CN little on DISTR-long |
| Language | Tsez |
| Transcript | Ža reƛ'izał maxšel, hunar yołäsi, saɣaw ci rik'äsi uži zown. |
| Seg. | ža reƛ'a-bi-ł maxšel hunar yoł-asi saɣaw ci r-ik'i-asi uži zow-n |
| Translation | He was a boy who was known for his talent and skill at his hands. |
| Gold Gloss | DEM1.SG hand-PL-CONT.ESS skill talent be-RES.PRT healthy name IV-go-RES.PRT boy be.NPRS-PST.UNW |
| GlossLM Gloss | DEM1.SG hand-PL-CONT.ESS talent talent be-RES.PRT common name IV-go-RES.PRT boy be.NPRS-PST.UNW |
| CWoMP (Train) Seg. | ža reƛ'a-bi-ł maxšel hunar yoł-asi saɣ ci r-ik'i-asi uži zow-n |
| CWoMP (Train) Gloss | DEM1.SG hand-PL-CONT.ESS skill talent be-RES.PRT cure name IV-go-RES.PRT boy be.NPRS-PST.UNW |
| CWoMP (Ext.) Seg. | ža reƛ'a-bi-ł maxšel hunar yoł-asi saɣaw ci r-ik'i-asi uži zow-n |
| CWoMP (Ext.) Gloss | DEM1.SG hand-PL-CONT.ESS skill talent be-RES.PRT healthy name IV-go-RES.PRT boy be.NPRS-PST.UNW |
| Language | Tsez |
| Transcript | Elur k'ukyoƛ ħažoƛin, ƛ'iri butin žedu. |
| Seg. | elu-r k'uk-ƛ ħaži-o-ƛin ƛ'iri b-uti-n žedu. |
| Translation | "Exchange it for a cap with us!", they proposed. |
| Gold Gloss | we(I).OBL-LAT cap-SUB.ESS exchange-IMPR-QUOT above I.PL-turn-PFV.CVB DEM1.IPL |
| GlossLM Gloss | we(I).OBL-LAT cap-SUB.ESS exchange-IMPR-QUOT above I.PL-return-PST.UNW DEM1.IPL |
| CWoMP (Train) Seg. | 'ža reƛ'a-bi-ł maxšel hunar yoł-asi saɣ ci r-ik'i-asi uži zow-n |
| CWoMP (Train) Gloss | we(I).OBL-LAT cap-SUB.ESS exchange-IMPR-QUOT above I.PL-return-PST.UNW DEM1.IPL |
| CWoMP (Ext.) Seg. | ža reƛ'a-bi-ł maxšel hunar yoł-asi saɣaw ci r-ik'i-asi uži zow-n |
| CWoMP (Ext.) Gloss | we(I).OBL-LAT cap-SUB.ESS exchange-IMPR-QUOT above I.PL-return-PST.UNW DEM1.IPL |

Table 13: Random selection of system outputs (Part 2 of 2)

| | |
|---|---|
| Language | Arapaho |
| Transcript | biikoo ceniibiikoo |
| Seg. | biikoo cenii-biikoo |
| Translation | Night , it was a really late at night . |
| Gold Gloss | at.night very-at.night |
| GlossLM Gloss | at.night very-at.night |
| CWoMP (Train) Seg. | biikoo cenii-biikoo |
| CWoMP (Train) Gloss | at.night very-at.night |
| CWoMP (Ext.) Seg. | biikoo cenii-biikoo |
| CWoMP (Ext.) Gloss | at.night very-at.night |
| Language | Arapaho |
| Transcript | ne'P honoo3itootowoo |
| Seg. | ne'-P honoo3itoot-owoo |
| Translation | I will tell it |
| Gold Gloss | that-pause IC.tell.story-1S |
| GlossLM Gloss | then-pause IC.tell.of-1S |
| CWoMP (Train) Seg. | ne'-P honoo3itoot-owoo |
| CWoMP (Train) Gloss | 'then-pause IC.tell.story-1S |
| CWoMP (Ext.) Seg. | tne'-P honoo3itoot-owoo |
| CWoMP (Ext.) Gloss | 'then-pause IC.tell.story-1S |
| Language | Nyangbo |
| Transcript | aábá anyɛ́núvɔɛ nɔ́ ɔbha |
| Seg. | a-á-bá a-nyɛ́núvɔɛ nɔ́ ɔ-bha |
| Gold Gloss | 3SG-PROG-come CM-boy DEF CM-side |
| GlossLM Gloss | 3SG-PROG-come CM-boy DEF CM-side |
| CWoMP (Train) Seg. | a-á-bá a-nyɛ́núvɔɛ nɔ́ ɔ-bha |
| CWoMP (Train) Gloss | 3SG-PROG-come CM-boy DEF CM-side |
| CWoMP (Ext.) Seg. | a-á-bá a-nyɛ́núvɔɛ nɔ́ ɔ-bha |
| CWoMP (Ext.) Gloss | 3SG-PROG-come CM-boy DEF CM-side |
| Language | Nyangbo |
| Transcript | anɛ bulĩ |
| Seg. | a-nɛ bu-lĩ |
| Gold Gloss | 3SG-NEG:be CM-water |
| GlossLM Gloss | 3SG-NEG:be CM-water |
| CWoMP (Train) Seg. | a-nɛ bu-lĩ |
| CWoMP (Train) Gloss | 3SG-NEG:be CM-water |
| CWoMP (Ext.) Seg. | a-nɛ bu-lĩ |
| CWoMP (Ext.) Gloss | 3SG-NEG:be CM-water |
| Language | Lezgi |
| Transcript | « Самур » газет атай береда . 00 . 20 |
| Seg. | « Самур » газет атун-й бере-е . 00 . 20 |
| Translation | When "Samur" newspaper came |
| Gold Gloss | « "Samur" » newspaper come-PTP time-INESS . 00 . 20 |
| GlossLM Gloss | Samur newspaper come-PTP time-INESS 00 20 |
| CWoMP (Train) Seg. | « Самур » газет атай-й бере-е . 00 . 20 |
| CWoMP (Train) Gloss | « "Samur" » newspaper coming-PTP time-INESS . 00 . 20 |
| CWoMP (Ext.) Seg. | « Самур » газет атай-й бере-е . 00 . 20 |
| CWoMP (Ext.) Gloss | « "Samur" » newspaper coming-PTP time-INESS . 00 . 20 |
| Language | Lezgi |
| Transcript | Писвал ийиз кIанзава хьир гададиз |
| Seg. | писвал-вал ийи-з кIан-зва хьир гада-ди-з . |
| Translation | He wants to do evil to the guy. |
| Gold Gloss | evil-PURP make-DAT want-IMPF even boy-DIR-DAT . |
| GlossLM Gloss | evil-PURP make-DAT want-IMPF even boy-DIR-DAT |
| CWoMP (Train) Seg. | 'писвал-вал ийи-з кIан-зва хьир гада-ди-з . |
| CWoMP (Train) Gloss | evil-PURP make-DAT want-IMPF even boy-DIR-DAT . |
| CWoMP (Ext.) Seg. | писвал-вал ийи-з кIан-зва хьир гада-ди-з . |
| CWoMP (Ext.) Gloss | evil-PURP make-DAT want-IMPF even boy-DIR-DAT . |